\documentclass[final]{cvpr}

\usepackage{times}
\usepackage{epsfig}
\usepackage{graphicx}
\usepackage{amsmath}
\usepackage{amssymb}
\usepackage[normalem]{ulem}

\usepackage{overpic}
\usepackage{color}
\usepackage{algorithm}
\usepackage{algpseudocode}
\usepackage{dsfont}
\usepackage{wrapfig}
\usepackage{snapshot}
\usepackage[utf8]{inputenc} %
\usepackage[T1]{fontenc}    %
\usepackage{url}            %
\usepackage{booktabs}       %
\usepackage{amsfonts}       %
\usepackage{nicefrac}       %
\usepackage{microtype}      %

\usepackage[pagebackref=true,breaklinks=true,colorlinks,bookmarks=false]{hyperref}

\usepackage{enumitem}
\usepackage{relsize}

\usepackage{subfigure} %
\usepackage{booktabs}
\usepackage{multirow}

\usepackage[resetlabels,labeled]{multibib}
\newcites{A}{Additional References}

\renewcommand{\paragraph}[1]{\vspace{.2em}\noindent\textbf{#1}}

\usepackage{color}
\definecolor{turquoise}{cmyk}{0.65,0,0.1,0.3}
\definecolor{purple}{rgb}{0.65,0,0.65}
\definecolor{dark_green}{rgb}{0, 0.5, 0}
\definecolor{orange}{rgb}{0.8, 0.6, 0.2}
\definecolor{red}{rgb}{0.8, 0.2, 0.2}
\definecolor{darkred}{rgb}{0.6, 0.1, 0.05}
\definecolor{blueish}{rgb}{0.0, 0.3, .6}
\definecolor{light_gray}{rgb}{0.7, 0.7, .7}
\definecolor{pink}{rgb}{1, 0, 1}
\definecolor{greyblue}{rgb}{0.25, 0.25, 1}

\newcommand{\revised}[1]{#1}

\usepackage{comment}
\specialcomment{DRAFT}{\color{red}}{\color{black}}

\newcommand{\Figure}[1]{Figure~\ref{fig:#1}}
\newcommand{\Eq}[1]{Equation~\ref{eq:#1}}
\newcommand{\eq}[1]{(\ref{eq:#1})}

\newcommand{\Section}[1]{Section~\ref{sec:#1}}

\newcommand{\Algorithm}[1]{Algorithm~\ref{alg:#1}}

\newcommand{\Table}[1]{Table~\ref{tbl:#1}}

\newcommand{\CIRCLE}[1]{\raisebox{.5pt}{\footnotesize \textcircled{\raisebox{-.6pt}{#1}}}}

\renewcommand{\comment}[1]{}

\newcommand{\bp}{\mathbf{p}}
\newcommand{\bP}{\mathbf{P}}

\newcommand{\bh}{\mathbf{h}}

\newcommand{\bx}[0]{\mathbf{x}}

\newcommand{\by}[0]{\mathbf{y}}
\newcommand{\calL}{\mathcal{L}}

\newcommand{\IR}{\mathds{R}}

\DeclareMathOperator*{\minimize}{\text{minimize}}

\newcommand{\softmax}{\text{softmax}}

\newcommand{\opt}{\text{Optimization step for }}

\newcommand{\loss}{\mathcal{L}}

\definecolor{orange}{rgb}{.8,0.3,0}
\definecolor{blue}{rgb}{0,0,0.6}

\definecolor{color1}{RGB}{0,199,1}
\definecolor{color2}{RGB}{224,43,28}

\newcommand{\image}{\mathcal{I}}
\newcommand{\heatmap}{\mathbf{h}}

\newcommand{\synthesistask}{{image reconstruction}}
\newcommand{\losstask}{\calL_\text{task}}

\newcommand{\detector}{\mathcal{H}}
\newcommand{\detectorpars}{\eta}
\newcommand{\extractor}{\mathcal{E}_K}
\newcommand{\sampler}{\mathcal{S}}
\newcommand{\task}{\mathcal{T}}
\newcommand{\taskpars}{\tau}
\newcommand{\ipatch}{k}
\newcommand{\patches}{\{\patch_\ipatch\}}
\newcommand{\patch}{\mathbf{P}}
\newcommand{\parameters}{\{\mathbf{x}_k\}}

\newcommand{\splatter}{\mathcal{G}}

\def\etal{\emph{et al}.}

\def\cvprPaperID{2167} %
\def\confYear{CVPR 2021}

\begin{document}
\title{MIST: Multiple Instance Spatial Transformer}
\author{
Baptiste Angles$^{1,*}$, \hspace{3pt}
Yuhe Jin$^{2,*}$, \hspace{3pt}
Simon Kornblith$^{3}$, \\
Andrea Tagliasacchi$^{3}$, \hspace{3pt}
Kwang Moo Yi$^{1,2}$
\\[1em]
$^1$University of Victoria, \hspace{3pt} 
$^2$University of British Columbia, \hspace{3pt} \\
$^3$Google Research, \hspace{3pt}
$^*$equal contributions
}

\maketitle

\begin{abstract}
We propose a deep network that can be trained to tackle image reconstruction and classification problems that involve detection of multiple object instances, \textit{without} any supervision regarding their whereabouts. 
The network learns to extract the most significant K patches, and feeds these patches to a task-specific network -- e.g., auto-encoder or classifier -- to solve a domain specific problem.
The challenge in training such a network is the non-differentiable top-$K$ selection process. 
To address this issue, we lift the training optimization problem by treating the result of top-$K$ selection as a slack variable, resulting in a simple, yet effective, multi-stage training. 
Our method is able to learn to detect recurring structures in the training dataset by learning to reconstruct images. 
It can also learn to localize structures when only knowledge on the occurrence of the object is provided, and in doing so it outperforms the state-of-the-art.
\end{abstract}
\section{Introduction}
\label{sec:intro}

Finding and processing multiple instances of characteristic entities in a scene is core to many computer vision applications, including object detection~\cite{Ren15,He17,Redmon17}, pedestrian detection~\cite{Dollar12b,Stewart16,Zhang18c}, and keypoint localization~\cite{Lowe04,Bay08}.
In traditional pipelines, it is common to localize entities by selecting the top-K responses in a heatmap and use their locations~\cite{Lowe04,Bay08,Felzenszwalb10}.
However, this type of approach does not provide a gradient with respect to the heatmap, and cannot be directly integrated into neural network-based systems.

To overcome this challenge, previous work proposed to use grids~\cite{Redmon16,He17,DeTone17b} to simplify the formulation by isolating each instance~\cite{Yi16b}, or to optimize over multiple branches~\cite{Ono18}.
While effective, these approaches require additional supervision to localize instances, and do not generalize well outside their intended application domain.
Other formulations, such as sequential attention~\cite{Ba15,Gregor15,Eslami16} and channel-wise approaches~\cite{Zhang18b} are problematic to apply when the number of instances of the same object is large, as we show later through experiments.

\begin{figure*}[t]
\begin{center}
\begin{overpic} 
    [width=1.0\linewidth]
    {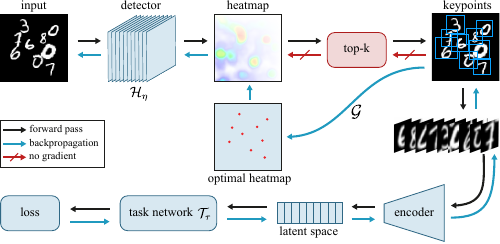}
    \put(88,28){\Large$\sampler$}
\end{overpic}
\end{center}
\vspace{-1.0em}
\caption{
\revised{
\textbf{The MIST architecture} -- 
A network $\detector_\detectorpars$ estimates locations and scales of patches encoded in a heatmap $\heatmap$.
Patches are then extracted via a sampler $\sampler$, and then fed to a task-specific network $\task_\taskpars$.
For example, the specific task could be to re-synthesize the image as a super-position of (unknown and locally supported) basis functions, or simply classifying each patch.
As top-k operation is non-differentiable, we back-propagate by lifting through $\splatter$; see \Section{optimization} for details.
}
}
\label{fig:arch}
\vspace{-1.0em}
\end{figure*}

Here, we introduce a novel way to tackle this problem, which we term \textit{Multiple Instance Spatial Transformer}, or \textit{MIST} for brevity. 
As illustrated in \Figure{arch} for the image synthesis task, given an image, we first compute a heatmap via a deep network whose local maxima correspond to locations of interest.
From this heatmap, we gather the parameters of the top-$K$ local maxima, and then extract the corresponding collection of image patches via an image sampling process.
We process each patch independently with a task-specific network, \eg, an image decoder, and aggregate the network's output across patches.

Training a pipeline that includes a non-differentiable selection/gather operation is non-trivial.
Thus, we propose to lift the problem to a higher dimensional one by treating the parameters 
defining the interest points as slack variables, and introduce a hard constraint that they must correspond to the output of the heatmap network.
This constraint is realized by introducing an auxiliary function that generates a heatmap given a set of interest point parameters.
We then solve for the relaxed version of this problem, where the hard constraint is turned into a soft one, and the slack variables are also optimized within the training process.
Critically, our training strategy allows the network to incorporate both non-maximum suppression and top-K selection.
We evaluate the performance of our approach for 
\CIRCLE{1} recovering the basis functions that created a given texture, 
\CIRCLE{2} detection and classification of handwritten digits in cluttered scenes, and 
\CIRCLE{3} object detection on natural images, all without any location supervision.

\paragraph{Contributions}
In summary, in this paper we:
\vspace{-0.4em}
\begin{itemize}[noitemsep,leftmargin=1.0em]
\item propose an end-to-end training method that allows the use of top-K selection;
\item show that our framework can reconstruct images as parts, as well as detect/classify instances without any location supervision;
\item outperform the state of the art in various scenarios, including on natural images.
\end{itemize}
\vspace{-0.3em}

\section{Related work}
\label{sec:related}
Focusing on a sub-region in an image is in an essence an attention model.
Attention models and the use of localized information have been actively investigated in the literature.
Some examples include discriminative tasks such as fine-grained classification~\cite{Sun18e}, pedestrian detection~\cite{Zhang18c}, and generative ones such as image synthesis from natural language~\cite{Johnson18}.
They have also been studied in the context of more traditional Multiple Instance Learning (MIL) setup~\cite{Ilse18}.
We now discuss a selection of representative works, and classify them according to how they deal with multiple instances.

\paragraph{Grid-based methods.}
Since the introduction of Region Proposal Networks (RPN)~\cite{Ren15}, grid-based strategies have been used for dense image captioning~\cite{Johnson16a}, instance segmentation~\cite{He17}, keypoint detection~\cite{Georgakis18}, and multi-instance object detection~\cite{Redmon17}.
Recent improvements to RPNs attempt to learn the concept of a generic object covering multiple classes~\cite{Singh18}, and to model multi-scale information~\cite{Chao18}.
The multiple transformation corresponding to separate instances can also be densely regressed via Instance Spatial Transformers~\cite{Wang18e}, which removes the need to identify discrete instance early in the network.
However, all these methods are fully supervised, requiring both class \textit{labels} and object \textit{locations} for training.

\paragraph{Heatmap-based methods.}
Heatmap-based methods have recently gained interest to detect features~\cite{Yi16b,Ono18,DeTone17b}, find landmarks~\cite{Zhang18b,Merget18}, and regress human body keypoints~\cite{Tekin17a,Newell16}.
While it is possible to output a separate heatmap per class~\cite{Zhang18b,Tekin17a}, most heatmap-based approaches do not distinguish between instances.
\cite{Yi16b} re-formulate the problem based on each instance, but in doing so introduce a sharp 
difference between training and testing.
Grids can also be used 
with heatmaps~\cite{DeTone17b}, but this results in an unrealistic 
assumption of uniformly distributed detections in the image.
Overall, heatmap-based methods excel when the ``final'' task of the network is to generate a heatmap~\cite{Merget18}, but are problematic to use as an intermediate layer in the presence of multiple instances.

\paragraph{Sequential inference methods.}
Another way to approach multi-instance problems is to attend to one instance at a time in a sequentially.
Training neural network-based models with sequential attention is challenging, but approaches using policy gradient~\cite{Ba15} and differentiable attention mechanisms~\cite{Gregor15,Eslami16} have achieved some success for images comprising \textit{small} numbers of instances.
However, Recurrent Neural Networks (RNN) often struggle to generalize to sequences longer than the ones encountered during training, and while recent results on inductive reasoning are promising~\cite{Gupta18}, their performance does not scale well when the number of instances is large.

\paragraph{Knowledge transfer.}
To overcome the acquisition cost of labelled training data, one can transfer knowledge from labeled to unlabeled dataset. 
For example, \cite{Inoue18} train on a single instance dataset, and then attempt to generalize to multi-instance domains, while \cite{Uijlings18} attempt to also transfer a multi-class proposal generator to the new domain.
While knowledge transfer can be effective, it is highly desirable to devise unsupervised methods such as ours that do not depend on an additional dataset.

\paragraph{Weakly supervised methods.}
To further reduce the labeling effort, weakly supervised methods have also been proposed.
\cite{Wan18} learn how to detect multiple instances of a single object via region proposals and ROI pooling, while \cite{Tang18} propose to use a hierarchical setup to refine their estimates.
\cite{Gao18} provides an additional supervision by specifying the number of instances in \textit{each} class, while \cite{Oquab15} and \cite{Zhang18f} localize objects by looking at the network activation maps~\cite{Zhou16e,Selvaraju17}.
\cite{shen2018generative} introduce an adversarial setup, where detection boxes are supervised by distribution assumptions and classification objectives.
Recently, \cite{Li_iccv19} proposed to accompany a segmentation task to improve the two-branch architecture of \cite{Bilen_cvpr16}.
However, all these methods still rely on region proposals from an existing method, or define them via a hand-tuned process.

\paragraph{Differentiable top-k methods.}
Recently, researchers proposed a differentiable formulation of top-K via optimal transport~\cite{Xie20}.
This method unfortunately often requires too much memory in its application, and is geared towards an image classification setup.
We show in \Section{res_classif} that our method performs better.

\section{MIST framework}
\label{sec:optimization}

We first explain our framework at a high level, and then detail each component in \Section{components}.
A prototypical MIST architecture (\Figure{arch}) is composed of two trainable components: 
\CIRCLE{1} the first module receives an image as input and extracts $K$ patches, at image locations and scales given by the top $K$ local maxima of a heatmap generated by a trainable heatmap network $\detector_\detectorpars$ with weights $\detectorpars$.
\CIRCLE{2} the second module processes each extracted patch with a task-specific network $\task_\taskpars$ whose weights $\taskpars$ are shared across patches, and further manipulates these signals to express a task-specific loss $\loss_{task}{}$.
The two modules are connected through non-maximum suppression on the scale-space heatmap output of $\detector_\detectorpars$, followed by a top-$K$ selection process to extract the parameters defining the patches, which we denote as $\extractor$.
We then sample patches at these locations through (differentiable) bilinear sampling $\sampler$ and feed them the task module.

The defining characteristic of the MIST architecture is that they are \textit{quasi-unsupervised}: the only supervision required is the number $K$ of patches to extract.
However, as we show in \Section{diffK}, our method is not sensitive to the choice of $K$ during training.
In addition, our extraction process uses a single heatmap for all instances that we extract.
In contrast, existing heatmap-based methods~\cite{Eslami16,Zhang18b} typically rely on heatmaps dedicated to \textit{each} instance, which is problematic when an image contains two instances of the same class.
Conversely, we restrict the role of the heatmap network~$\detector_\detectorpars$ to find the ``important'' areas in a given image, without having to distinguishing between classes, hence simplifying learning.

Formally, the training of the MIST architecture is summarized by:
\begin{align}
\minimize_{\taskpars,\detectorpars} \:\: & \loss_\text{task}(\task_\taskpars( \sampler(\extractor(\detector_\detectorpars(\image)))))
\label{eq:misttrain}
\end{align}
where $\taskpars,\detectorpars$ are the network trainable parameters. 
Unfortunately, the patch extractor $\extractor$ is non-differentiable, as it identifies the locations of the top-$K$ local maxima of a heatmap and then selects the corresponding patches from the input image.
Differentiating this operation provides a gradient with respect to the input, but no gradient with respect to the heatmap.
Although it is possible to smoothly relax the patch selection operation in the $K{=}1$ case~\cite{Yi16b} (\ie, $\mathrm{argmax}$), it is unclear how this can be generalized to the case of \textit{multiple} distinct local maxima.
It is thus impossible to train the patch selector parameters directly by backpropagation.
Here, we propose an alternative 
via lifting.

\begin{algorithm*}[t]
\caption{
Multi-stage optimization for MISTs
}
\label{alg:training}
\begin{algorithmic}[1]
    \Require{
      $K$ : number of patches to extract,
      $\losstask$ : task specific loss,
      $\image$ : input image,
      $\splatter$ : the keypoints to heatmap function,
      $\detector$ : the heatmap network,
      $\detectorpars$ : parameters of the heatmap network,
      $\task$ : the task network,
      $\taskpars$ : parameters of the task network,
      $\extractor$ : the top-$K$ operator.
    }
    \Function{TrainMIST}{$\image$, $\losstask$}
    \For{each training batch}
    \State{$\taskpars \leftarrow \opt \task_\taskpars$ with $\losstask + \lambda\|\parameters -  \extractor(\detector_\detectorpars(\image)) \|_2^2$}
    \State{$\parameters \leftarrow \opt \parameters$ with $\losstask + \lambda\|\parameters -  \extractor(\detector_\detectorpars(\image)) \|_2^2$}
    \State{$\detectorpars \leftarrow \opt \detectorpars$ with $\|\splatter(\parameters) -  \detector_\detectorpars(\image) \|_2^2$}
    \EndFor
    \EndFunction
    \end{algorithmic}
\end{algorithm*}

\paragraph{Differentiable top-K via lifting.}
The introduction of auxiliary variables to simplify the structure of an optimization problem has proven effective in a range of domains ranging from non-rigid registration~\cite{Taylor16}, to robust optimization~\cite{Zach18}.
To simplify our training optimization, we start by decoupling the heatmap tensor from the optimization~\eq{misttrain} by introducing the corresponding auxiliary variables~$\heatmap$, as well as the patch location variables~$\parameters$ from the top-K extractor:
\begin{align}
\minimize_{\detectorpars,\taskpars,\heatmap,\parameters} \:\: & \loss_\mathrm{task}(\task_\taskpars( \sampler(\parameters)))
\label{eq:con_det}\\
\mathrm{\quad s.t.\quad } &
\heatmap = \detector_\detectorpars(\image), 
\quad
\parameters = \extractor(\heatmap).\nonumber
\end{align}
\revised{
We then relax the first constraint above to a least-squares penalty via a Lagrange multiplier $\lambda$:
\begin{align}
\minimize_{\detectorpars,\taskpars,\heatmap,\parameters} \:\: & \loss_\mathrm{task}(\task_\taskpars( \sampler(\parameters))) + \lambda\|\heatmap -  \detector_\detectorpars(\image) \|_2^2
\label{eq:energy_2}\\
\mathrm{\quad s.t.\quad } & \parameters = \extractor(\heatmap).\nonumber
\end{align}
As in many methods that use keypoint supervision to regress heatmaps~\cite{cao2018openpose}, we assume that a good heatmap generator $\splatter$ exists -- that is $\parameters \approx\extractor(\splatter(\parameters))$
We can now rewrite our optimization as:
\begin{align}
\minimize_{\detectorpars,\taskpars,\heatmap,\parameters} \:\: & \loss_\mathrm{task}(\task_\taskpars( \sampler(\parameters))) + \lambda\|\heatmap -  \detector_\detectorpars(\image) \|_2^2\\
\mathrm{\quad s.t.\quad } &
\heatmap = \splatter(\parameters). \nonumber
\end{align}
We can now drop the auxiliary variable $\heatmap$ and rewrite our optimization as:
\begin{align}
\minimize_{\detectorpars,\taskpars,\parameters} \:\: \loss_\mathrm{task}&(\task_\taskpars( \sampler(\parameters))) + 
\label{eq:energy_1}\\ 
&\lambda\|\splatter(\parameters) -  \detector_\detectorpars(\image) \|_2^2,  \nonumber
\end{align}
and then approach the problem by \textit{block coordinate descent} -- where the energy terms not containing the variable being optimized are safely dropped, and we apply $\extractor$ to the penalty term of \eq{energy_1}:
\begin{align}
\minimize_{\taskpars,\parameters} \:\:  \loss_\mathrm{task}&(\task_\taskpars( \sampler(\parameters)))
+ \label{eq:phase1}\\
&\lambda
\|\parameters -  \extractor(\detector_\detectorpars(\image)) \|_2^2,\nonumber
\end{align}
\begin{align}
\minimize_{\detectorpars} \:\: & \|\splatter(\parameters) -  \detector_\detectorpars(\image) \|_2^2.
\label{eq:phase2}
\end{align}
}
To accelerate training, we further split \eq{phase1} into two stages, and alternate between optimizing $\taskpars$ and $\parameters$.
Being based on block coordinate descent, this process converges smoothly, as we show in Section~\ref{sec:convergence}.
The summary for the three stage optimization procedure is outlined in \Algorithm{training}.
Notice that we are not introducing \textit{any} additional supervision signal that is tangent to the given task.

\section{Implementation}
\label{sec:components}

\subsection{Components}
\label{sec:architecture}

\paragraph{Multiscale heatmap network -- $\detector_\detectorpars$.}
Any network that provides localization via a heatmap can be used. 
Our standard implementation is a multiscale heatmap network inspired by LF-Net~\cite{Ono18}.
We improve the network by applying modifications on how the scores are aggregated over scale. 
The network takes as input an image $\image$, and outputs multiple heatmaps $\bh'_s$ of the same size for each scale level $s$.
To limit the number of necessary scales, we use a discrete scale space with $S$ scales, and resolve intermediate scales via 
interpolation.
For tasks where pre-trained networks are needed -- \eg classification on a natural image -- we utilize an architecture similar to~\cite{He17}.
Note that we are simply using the pretrained deep features, and these can also be obtained through a fully unsupervised setup, such as contrastive learning~\cite{Chen20e}, if desired; see~\Section{heatmap_net} of the appendix for details.

\paragraph{Top-K patch selection -- $\extractor$.} 
To extract the top $K$ elements, we perform an addition cleanup through a standard non-maximum suppression.
We then find the spatial locations of the top $K$ elements of this suppressed heatmap, denoting the spatial location of the $k$\textsuperscript{th} element as $(x_k, y_k)$, which now should correspond to local maxima.
When using multiscale heatmaps, for each location, we also compute the corresponding scale by weighted first order moments~\cite{Suwajanakorn18a} where the weights are the responses in the corresponding heatmaps, \textit{i.e.} $s_k = \sum_s \bh'_s(x_k, y_k) s$

\paragraph{Generative model for ideal heatmap -- $\splatter(\parameters)$.}
For the generative model that turns keypoint and patch locations into heatmaps, we apply a simple model where the heatmap is zero everywhere except at the corresponding keypoint locations (patch centers); see~\Section{inversemap} of the appendix.

\paragraph{Patch resampling -- $\sampler$.}
As a patch is uniquely parameterized by its location and scale, i.e. $\bx_k\!=\!(x_k, y_k, s_k)$, we can then proceed to resample its corresponding tensor via bilinear interpolation~\cite{Jaderberg15,wei_iccv19} as~$\patches~=~\sampler \left(\image, \parameters \right)$.

\subsection{Task-specific networks}
\label{sec:application}
We now introduce two applications of the MIST framework.
We use the \textit{same} heatmap network and patch extractor for both applications, but the task-specific network and loss differ.
We provide further details regarding the task-specific network architectures in Section~\ref{sec:implementation} of the appendix.

\paragraph{Image reconstruction / auto-encoding.}
\label{sec:synthesis}
As illustrated in \Figure{arch}, for \synthesistask{} we append our patch extraction network with a \textit{shared} auto-encoder for each extracted patch.
We can then train this network to \textit{reconstruct} the original image by inverting the patch extraction process and minimizing the mean squared error between the input and the reconstructed image. %
Overall, the network is designed to \textit{jointly} model and localize repeating structures in the input signal.
Specifically, we introduce the generalized inverse sampling operation~$\sampler^{-1}(\bP_i, \bx_i)$, which starts with an image of all zeros, and places the patch $\bP_i$ at $\bx_i$.
We then sum all the images together to obtain the reconstructed image, optimizing the task loss
\begin{equation}
    \loss_\mathrm{task} = \mathlarger{\mathlarger{\|}}\image - \sum_{i}\sampler^{-1}\left(\bP_i, \bx_i\right)\mathlarger{\mathlarger{\|}}_2^2.
\label{eq:unsup}
\end{equation}

\begin{figure*}
\centering
\begin{overpic} 
    [width=0.98\linewidth]
    {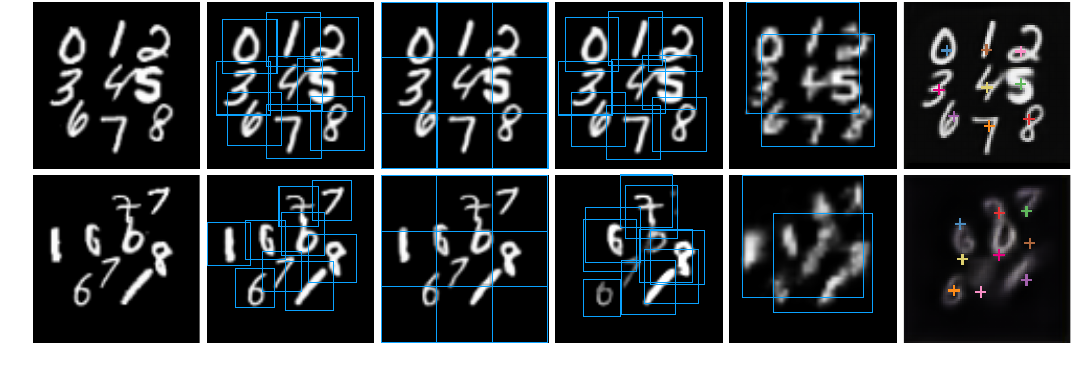}
    \put(6.9,1){\small{input image}}
    \put(25,1){\small{MIST}}
    \put(42,1){\small{grid}}
    \put(55,1){\small{channel-wise}}
    \put(72,1){\small{Eslami et al.}}
    \put(88,1){\small{Zhang et al.}}
    \put(1,22){\rotatebox{90}{\small{MNIST easy}}}
    \put(1,6){\rotatebox{90}{\small{MNIST hard}}}
\end{overpic}
\caption{
MNIST character synthesis examples for (top) the ``easy'' single instance setup and (bottom) the hard multi-instance setup.
We compare the output of MISTs to grid, channel-wise, sequential Eslami~\etal~\cite{Eslami16} and Zhang~\etal~\cite{Zhang18b}.
}
\label{fig:mnist_gen}
\vspace{-0.5em}
\end{figure*}

\label{sec:classification}
\paragraph{Multiple instance classification.}
By appending a classification network to the patch extraction module, we can also perform multiple instance learning.
For each extracted patch $\patch_k$ we apply a shared classifier network to output $\hat{\by}_k \in \IR^{C}$, where $C$ is the number of classes.
In turn, these are then converted into probability estimates by the transformation $\hat{\bp}_k = \softmax(\hat{\by}_k)$.

With $\by_{l}$
being the one-hot ground-truth labels of instance $l$ with unit norm, we define the multi-instance classification loss as
\begin{equation}
    \loss_\mathrm{task} = \left\|
        \frac{1}{L}\sum_{l=1}^{L}\by_l - \frac{1}{K}\sum_{k=1}^{K}\hat{\bp}_k
    \right\|_2^2,
\label{eq:mil_easy}
\end{equation}
where 
$L$ is the number of instances in the image.
We empirically found this loss to be more stable compared to the cross-entropy loss, similar to \cite{Mao17}.
Note that we \textit{do not} provide supervision about the localization of instances, yet the detector network will automatically learn how to localize with minimal supervision (\ie, the number of instances).

\section{Results and evaluation}
\label{sec:results}
To demonstrate the effectiveness of our framework we apply MIST to two different tasks.
We first perform a quasi-unsupervised image reconstruction task, where \emph{only} the total number of instances in the scene is provided, \ie, K is defined.
We then show that MIST can also be applied to weakly supervised multi-instance classification, where only image-level supervision is provided.
Note that, unlike region proposal based methods, our method relies only on the cues from the classifier, do not require object proposals, and can be trained from scratch.

\subsection{Image reconstruction}
From the MNIST dataset, we derive two different scenarios.
In the \textit{MNIST easy} dataset, we consider a simple setup where the \emph{sorted} digits are confined to a perturbed \emph{grid} layout; see \Figure{mnist_gen}~(top).
Specifically, we perturb the digits with a Gaussian noise centered at each grid center, with a standard deviation that is equal to one-eighths of the grid width/height.
In the \textit{MNIST hard} dataset, the positions are randomized through a Poisson distribution~\cite{Bridson07}, as is the identity, and cardinality of each digit.
We allow multiple instances of the same digit to appear in this variant.
For both datasets, we construct both training and test sets, and the test set is never seen during training.

\paragraph{Comparison baselines.}
We compare our method against four baselines \textbf{\CIRCLE{1}} the \emph{grid} %
method divides the image into a $3\times3$ grid
and applies the same auto-encoder architecture as MIST to each grid location to reconstruct the input image;
\textbf{\CIRCLE{2}} the \emph{channel-wise} method uses the same auto-encoder network as MIST, but we modify the heatmap network to produce $K$ channels as output, where each channel is dedicated to an interest point.
Locations are obtained through a channel-wise soft-argmax as in \cite{Zhang18b};
\textbf{\CIRCLE{3}} {\it Esl16}~\cite{Eslami16} is a sequential generative model;
\textbf{\CIRCLE{4}} {\it Zha18}~\cite{Zhang18b} is a 
heatmap-based method with channel-wise strategy for unsupervised learning of landmarks. 
We do not compare against Xie20~\cite{Xie20} for the reconstruction tasks as it requires too much memory to be used for generative tasks;
see \Section{baseline_details} of the appendix for more details.

\paragraph{Results for ``MNIST easy''.}
As shown in \Figure{mnist_gen}~(top) all methods successfully re-synthesize the image, with the exception of Eslami~et~al.~\cite{Eslami16}. 
As this method is sequential, with nine digits the sequential implementation simply becomes too difficult to optimize through.
Note how this method only learns to describe the scene with a few large regions.
We summarize quantitative results in \Table{recon}.

\begin{table}
\begin{center}
\resizebox{\linewidth}{!}{
\begin{tabular}{c c c c c c }
    \toprule
    & MIST & Grid & Ch.-wise & Esl16~\cite{Eslami16} & Zha18~\cite{Zhang18b} \\
    \midrule
    {MNIST easy} & {\bf .038} & .039 & .042 & .100 & .169 \\
    MNIST hard & {\bf .053} & .062 & .128 & .154 & .191 \\
    Gabor & {\bf.095} & - & - & - & - \\
    \bottomrule
\end{tabular}
}
\end{center}
\caption{Reconstruction error (root mean square error). 
Note that Grid \textit{does not} learn any notion of digits.
}
\label{tbl:recon}
\vspace{-1em}
\end{table}

\begin{figure*}
\centering
\begin{overpic} 
    [width=0.98\linewidth]
    {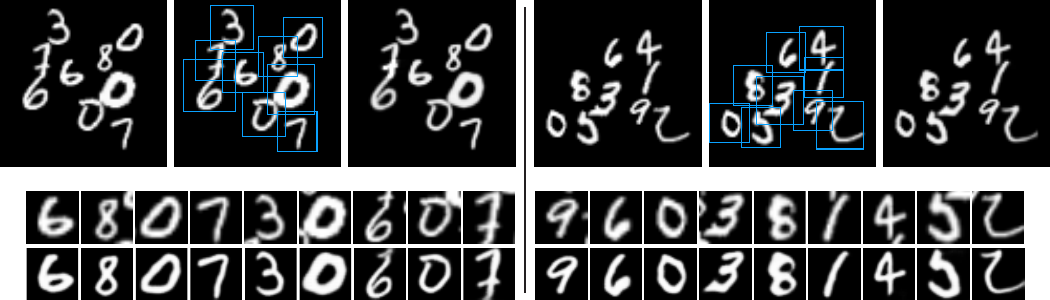}
    \put(3,11){\small{input image}}
    \put(20,11){\small{detections}}
    \put(37,11){\small{synthesis}}
    \put(54,11){\small{input image}}
    \put(71,11){\small{detections}}
    \put(88,11){\small{synthesis}}
    \put(0,0){\rotatebox{90}{\small{auto-encoder}}}
    \put(98,0){\rotatebox{90}{\small{auto-encoder}}}
\end{overpic}
\caption{
Two auto-encoding examples learnt from MNIST-hard.
In the top row, for each example we visualize input, patch detections, and synthesis.
In the bottom row we visualize each of the extracted patch, and how it is modified by the learnt auto-encoder.
}
\label{fig:mnist_gen_us}
\end{figure*}
\def \gaborsize {0.245}
\begin{figure*}
\centering
\includegraphics[width=0.98\linewidth]{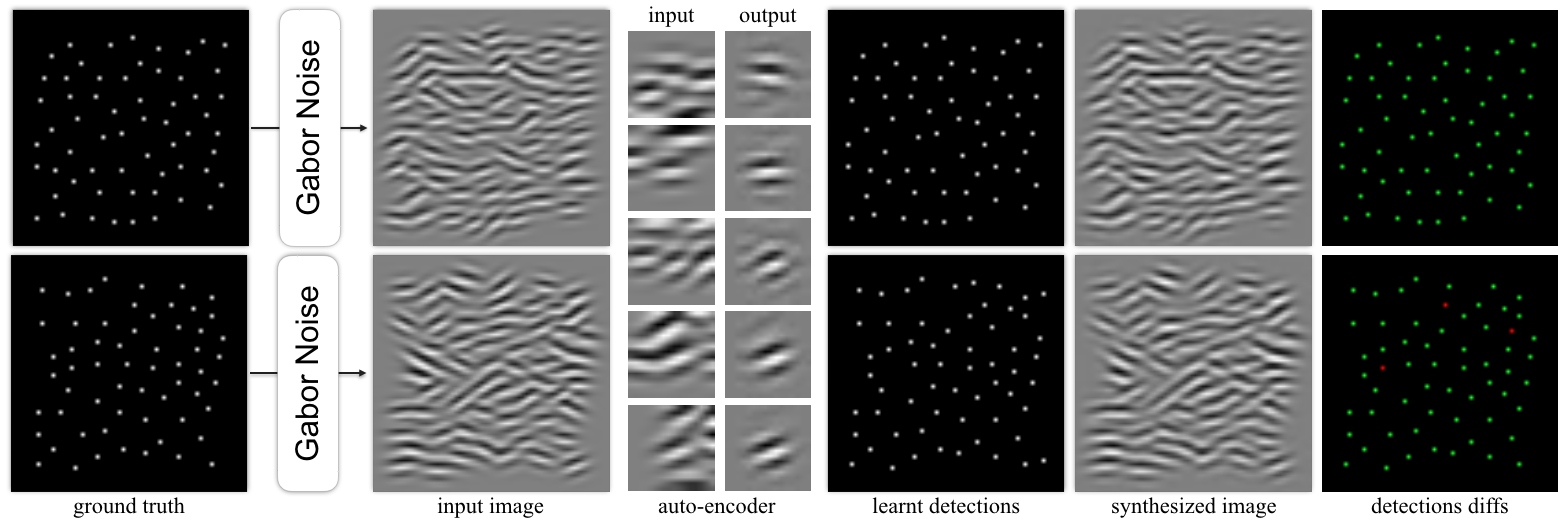}
\caption{
Inverse rendering of Gabor noise; we annotate {\color{dark_green}\textbf{correct}} / {\color{red} \textbf{erroneous}} localizations.
}
\vspace{-1em}
\label{fig:gabor}
\end{figure*}

\paragraph{Results for ``MNIST hard''.}
As shown in \Figure{mnist_gen}~(bottom), all baseline methods failed to properly represent the image. 
Only MIST succeeded at both localizing digits and reconstructing the original image. %
Although the grid method accurately reconstructs the image, it has no concept of individual digits.
Conversely, as shown in \Figure{mnist_gen_us}, our method generates accurate bounding boxes for digits even when these digits overlap, and does so without any location supervision. %
For quantitative results, please see~\Table{recon}.

\paragraph{Finding the basis of a procedural texture.}
We further demonstrate that our methods can be used to find the basis function of a procedural texture.
For this experiment we synthesize textures with procedural Gabor noise~\cite{gabor}.
Gabor noise is obtained by convolving oriented Gabor wavelets with a Poisson impulse process.
Hence, given exemplars of noise, our framework is tasked to regress the underlying impulse process and reconstruct the Gabor kernels so that when the two are convolved, we can reconstruct the original image.
\Figure{gabor} illustrates the results of our experiment.
The auto-encoder learned to accurately reconstruct the Gabor kernels, even though in the training images they are heavily overlapped.
These results show that MIST is capable of generating and reconstructing large numbers of instances per image, which is  \textit{intractable} with other approaches.

\def \mnistdfig {0.16}
\begin{figure*}
\centering
\includegraphics[width=\linewidth]{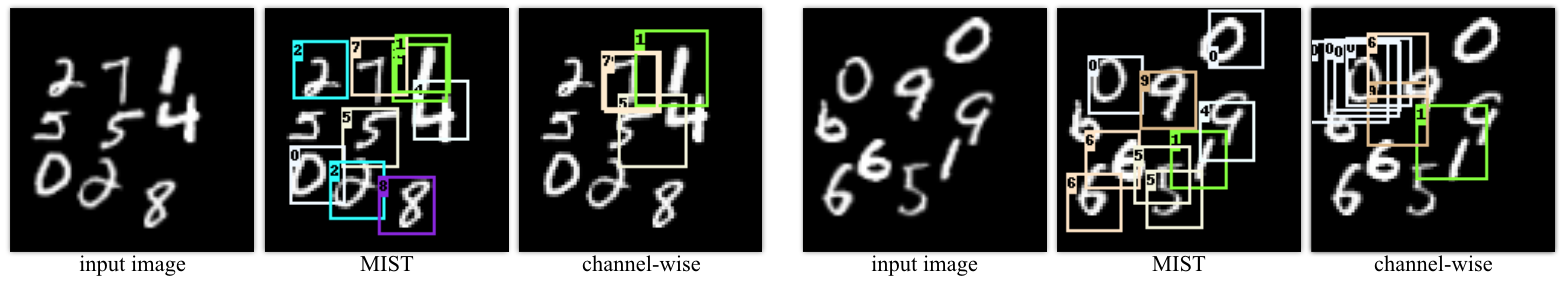}
\vspace{-1.2em}
\caption{
Two qualitative examples for detection and classification on our 
\textit{MNIST-hard}
dataset.
}
\label{fig:mnist_discrim}
\end{figure*}

\subsection{Multiple instance classification}
\label{sec:res_classif}

\paragraph{Multi-MNIST -- \Figure{mnist_discrim}, \Table{classif}, and \Table{classif_difftopk}.}
To test our method in a multiple instance classification setup, we rely on the \textit{MNIST hard} dataset.
To evaluate the detection accuracy of the models, we compute the intersection over union (IoU) between the ground-truth bounding box and the detection results, and assign it as a match if the IoU score is over~50\%.
In \Table{classif}, we compare our method to \textit{channel-wise} and report the number of correctly classified matches, as well as the proportion of instances that are both correctly detected and correctly classified.
To provide a sense of the upper-bound, we also provide results for the case when location supervision is used.
Our method clearly outperforms the \textit{channel-wise} strategy.
Note that, even without localization supervision, our method correctly localizes digits.
Conversely, the \textit{channel-wise} strategy fails to learn.
This is because \emph{multiple instances} of the same digits are present in the image.
For example, in the  \Figure{mnist_discrim}~(right), we have three sixes, two zeros, and two nines. 
This prevents any of these digits from being detected/classified properly by a channel-wise approach.

We further compare our method to Xie20~\cite{Xie20}, a state-of-the-art differentiable top-K formulation in \Table{classif_difftopk}.
We test with the \textit{MNIST hard} dataset, and also an easier version of this dataset with only three digits.
For Xie20~\cite{Xie20}, we modify the heatmap network to accommodate for the fact that Xie20 requires extracting \emph{all} potential patches, thus needing extensive memory, to be used in a detection framework (Xie20 generates a differentiable top K ``mask'', not a selection, which has then to be multiplied to the input to finally perform selection).
Hence, we apply an architecture similar to the region proposal network~\cite{He17}, which we found to work better than training a heatmap network that downsamples the heatmap to a reasonable size; see Supplementary \Section{implementation} for details.
Our method works best for both experiments in \Table{classif_difftopk}, with the gap between MIST and Xie20 widening \textit{significantly} as the problem becomes more complicated.

\begin{table}
\begin{center}
\resizebox{.8\linewidth}{!}{
\begin{tabular}{c c c|c}
    \toprule
    & MIST & \multicolumn{1}{c}{Ch.-wise} & \multicolumn{1}{c}{Supervised} \\
    \midrule
    IOU 50\% & 97.8\% & 25.4\% & 99.6\% \\
    Classif. & 98.8\% & 75.5\% & 98.7\%\\
    Both & \bf{97.5\%} & 24.8\% & 98.6\%\\
    \bottomrule
\end{tabular}
}
\end{center}
\caption{
Instance detection and classification on 
\textit{MNIST-hard}.
\label{tbl:classif}
}
\end{table}

\begin{figure*}
    \centering
    \includegraphics[width=\linewidth]{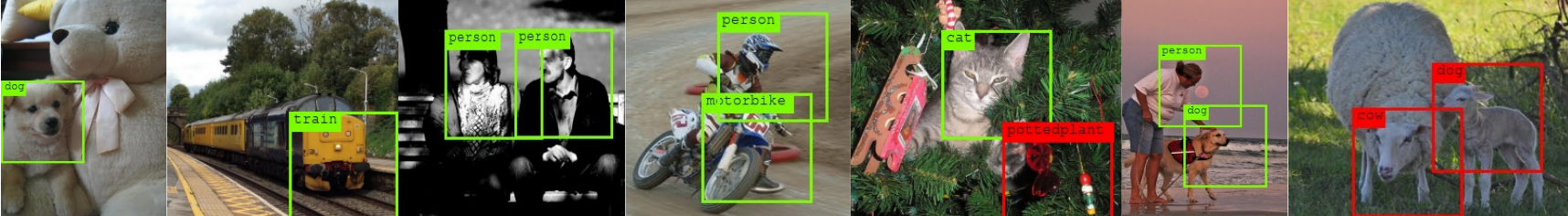}
    \includegraphics[width=\linewidth]{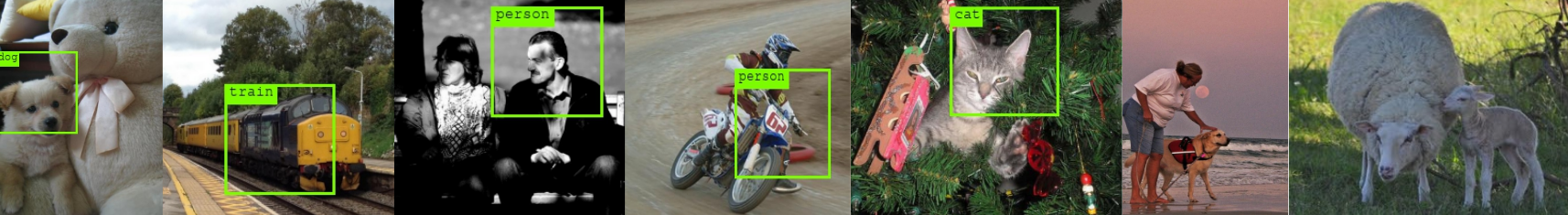}

    \caption{
    Qualitative results on PASCAL+COCO dataset.
    (Top row) we show detection results when using MIST and (bottom row) when using Xie20~\cite{Xie20}.
    Correctly localized and classified ones are shown in \textcolor{green}{\textbf{green}}, and wrong ones in \textcolor{red}{\textbf{red}}.
    }
    \label{fig:pascal_coco}
\end{figure*}

\paragraph{PASCAL + COCO -- \Table{pascal_coco} and \Figure{pascal_coco}.}
To further test MIST in a more generalized setting, we apply MIST on PASCAL VOC 2007 dataset~\cite{Everingham10} and COCO dataset~\cite{Lin14a}.
We combine the two datasets together by taking images from both datasets that contain any of the twenty Pascal VOC 2007 classes.
As we are aiming for an ambitious goal of not having any location supervision at all, including the bounding box proposals used in state-of-the-art weakly-supervised works~\cite{Bilen_cvpr16,Peng_cvpr17,Li_iccv19,Wei_eccv18}, we apply additional treatments to make the task easier.
We filter the combined dataset by removing images that contain objects that are too large (covering more than 30\% of the image) and images with objects that are too wide or narrow (aspect ratio bigger than 2.0 or smaller than 0.5 ).
The remaining images are random cropped to ensure that all objects within the image cover 30\% of the image in average.
The final dataset has 9780 images in total, which we split into training (7816 images), validation (987 images), and test (977 images) sets.

\begin{table}
\begin{center}
\resizebox{\linewidth}{!}{
\begin{tabular}{c c c|c c}
    \toprule
    & \multicolumn{2}{c}{\textit{MNIST-3 digits}} & \multicolumn{2}{c}{\textit{MNIST-hard}} \\
    & MIST & \multicolumn{1}{c}{Xie20~\cite{Xie20}} & MIST & \multicolumn{1}{c}{Xie20~\cite{Xie20}} \\
    \midrule
    IOU 50\% & {99.5}\% & {95.4}\% & {97.8}\% & {72.7}\%  \\
    Classif. & {96.9}\% & {97.4}\% & {98.8}\% & {93.1}\%  \\
    Both & {\textbf{96.6}}\% & {92.2}\% & {\textbf{97.5}}\% & {71.3}\%  \\
    \bottomrule
\end{tabular}
}
\end{center}
\caption{
Instance level detection and classification performance comparison with a state-of-the-art differentiable top K formulation~\cite{Xie20}.
\label{tbl:classif_difftopk}
}
\end{table}

As the task is hard, we further restrict the role of MIST and Xie20~\cite{Xie20} to \textit{localization} only without scale, \ie, we use a single scale heatmap for localization.
We use $K{=}2$ for this dataset.
The number of objects vary from one image to another in this dataset, thus, on top of the 20 classes in Pascal VOC 2007, we also add an additional class `background' to indicate unused bounding boxes.
With this setup, $K{=}2$ serves as the upper bound for number of instances to attend to in an image.

Since the dataset is relatively limited, as mentioned earlier in \Section{architecture}, we use a pretrained ResNet34~\cite{He16} network.
We use the feature maps of the fourth convolutional block and append a $1\times1$ convolution layer on top to obtain the heatmap.
We also resample patches from this heatmap instead of the original image to take advantage of these pretrained features.
We further apply data augmentation by performing random horizontal flips, as well as randomly masking the input image with a square mask of uniform distributed random size (between $0.15\times$ and $0.6\times$ of the image size) and filling the inside of the mask with the average color within the mask.
We emphasize that even with such filtering and restriction, our method is, to the best of our knowledge, the \textit{first} to be able to learn to detect and classify without any location supervision -- existing weakly supervised methods~\cite{Bilen_cvpr16,Peng_cvpr17,Li_iccv19,Wei_eccv18} mainly rely on provided object proposals.
Extending MIST to be able to deal with scale and aspect-ratio for natural images is left as future work.

We evaluate the performance of each method by computing the accuracy, recall, and F1-score of the detection outcomes. 
Regarding the localization, as we estimate the center only (scale is fixed), we consider the detection to be correct if the detection center is within the ground-truth bounding box.
We report results for MIST and Xie20~\cite{Xie20} in \Table{pascal_coco}, and show qualitative highlights in \Figure{pascal_coco}.
Our method outperforms Xie20~\cite{Xie20} by more than \textbf{13\% relative}, in terms of the F1-Score, for this task.

\begin{table}
\begin{center}
\resizebox{0.8\linewidth}{!}{
\begin{tabular}{l c c c}
    \toprule
    & Precision & Recall & F1-Score  \\
    \midrule
    MIST & \textbf{87.4}\% & \textbf{55.5}\% & \textbf{67.9}\% \\
    Xie20~\cite{Xie20} & 72.7\% & 51.0\% & 60.0\% \\
    \bottomrule
\end{tabular}
}
\end{center}
\caption{
Instance level localization and classification results on PASCAL+COCO dataset. 
\label{tbl:pascal_coco}
}
\end{table}

\subsection{Additional results}

\begin{figure}
\centering
\subfigure[$\loss_\mathrm{task}$]{
    \includegraphics[width=0.45\linewidth]{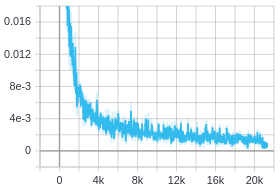}
}
\subfigure[$\|\splatter(\parameters) -  \detector_\detectorpars(\image) \|_2^2$]{
    \includegraphics[width=0.46\linewidth, ]{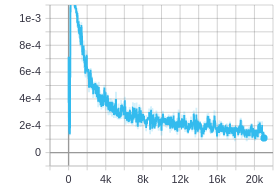}
}
\caption{
Evolution of the loss and heatmap during training on \textit{MNIST hard}.
(a) the task (classification) loss $\loss_\mathrm{task}$ and (b) the heatmap related loss $\|\splatter(\parameters) -  \detector_\detectorpars(\image) \|_2^2$ for each iteration.
 }
\label{fig:opt_appendix}
\end{figure}

\paragraph{Convergence during training.}
\label{sec:convergence}
As is typical for neural network training, our objective is non-convex and there is no guarantee that a local minimum found by gradient descent training is a global minimum. 
Empirically, however, the optimization process is stable, as shown in \Figure{opt_appendix}.
We report both the training losses $\loss_\mathrm{task}$ and $\|\splatter(\parameters) -  \detector_\detectorpars(\image) \|_2^2$ 
when training for \textit{MNIST hard}.
Both losses converge smoothly, demonstrating the stability of our formulation.
This is unsurprising, as our formulation is a lifted version of this loss to allow gradient-based training.
Note that the initial jump in \Figure{opt_appendix}~(b) is caused by the weak response of the randomly initialized detector's weights, producing a heatmap close to zero for every pixel.

\begin{table}
\begin{center}
    
\resizebox{0.6\linewidth}{!}{
\begin{tabular}{c c}
    \toprule
    Instances during train & AP$^\text{IoU=.50}$ \\
    \midrule
    $\{\mathbf{9}\}$ & \bf{92.2\%} \\
    $\{6,\mathbf{7},8,9\}$ & 90.1\% \\
    $\{3,4,5,\mathbf{6},7,8,9\}$ & 90.8\% \\
    \bottomrule
\end{tabular}
}
\end{center}
\caption{
Sensitivity experiment of $K$ on the MNIST hard dataset. 
We mark the number of $K$ used for testing in bold.
\label{tbl:k-table}
}
\end{table}
 
\revised{
\paragraph{Sensitivity to $K$ -- \Table{k-table}.}
\label{sec:diffK}
To investigate the sensitivity to the correctness of $K$ during training, we train with varying number of of instances and test with ground-truth $K$.
For example, with $K=6$, where the ground truth could be anything within $\{3,4,5,6,7,8,9\}$ -- we mark the $K$ used during training with bold.
Our method still is able to give accurate result with inaccurate $K$ during training.
Knowing the exact number of objects is not a strict requirement at test time, as our detector generates a heatmap for the entire image
regardless of the $K$ it was trained with.
Note that while in theory sequential methods are free from this limitation, in practice they are able to deal with limited number of objects (e.g. up to three) due to their recurrent nature.
}

\section{Conclusion}
We have introduced the MIST framework for multi-instance image reconstruction/classification.
With MIST, we show how localization of multiple instances can be learned even with the non-differentiable top-K operation by lifting.
We demonstrated the effectiveness of MIST by 
showing its compelling performance in both reconstruction and classification tasks using synthetic data.
We further demonstrated its performance in learning to localize and classify multiple objects in real-world images, without any supervision on location, and without any help from object proposal methods.
MISTs are a first step towards the definition of optimizable image decomposition networks that could be extended to a number of exciting \emph{unsupervised} learning tasks.

\section*{Acknowledgements}
This work was supported by the Natural Sciences and Engineering Research Council of Canada (NSERC) Discovery Grant, NSERC Collaborative Research and Development Grant, Google, and by Compute Canada.
We would also like to thank Shahram Izadi for his great support for this project.

{\small
\bibliographystyle{ieee_fullname}
\bibliography{string,vision,learning,add}
}

\clearpage

\clearpage
\renewcommand{\paragraph}[1]{\vspace{1em}\noindent\textbf{#1}}

\twocolumn[
\centering
\Large
\textbf{MIST: Multiple Instance Spatial Transformer} \\
\vspace{0.5em}Supplementary Material \\
\vspace{1.0em}
]
\appendix

\section{Heatmap network 
architectures
}
\label{sec:heatmap_net}

\vspace{-1em}
\paragraph{Standard architecture.}
Our 
standard architecture is the
multiscale heatmap network is inspired by LF-Net~\cite{Ono18}.
We employ a fully convolutional network to produce a single heatmap for each scale indexed by $s=1 \dots S$, on the input image~$\image$.
Specifically, for each scale we first scale the image proportionally to the inverse of the scale producing~$\image_s$, execute the network $\detector_\detectorpars$ on it, and finally scale the heatmap back to the original resolution.
This strategy ensures that the network cannot implicitly favor a particular scale and produces scale-independent responses.

This process generates a multiscale heatmap tensor $\heatmap = \{\heatmap_s\}$ of size $H\times W\times S$ where $\heatmap_s=\detector_\detectorpars(\image_s)$, and $H$ is the height of the image and $W$ is the width.
For the convolutional network we use $4$ ResNet blocks~\cite{He15}, where each block is composed of two $3\times3$ convolutions with $32$ channels and relu activations without any downsampling.
We then perform a \textit{local spatial softmax} operator \cite{Ono18} with spatial extent of $15\times15$ to sharpen the responses.
Then we further relate the scores across different scales by performing a ``softmax pooling'' operation over scale.
Specifically, if we denote the heatmap tensor after local spatial softmax as {\small $\tilde{\bh} = \{ \tilde{\bh}_s \} $}, since after the local spatial softmax {\small $\detector_\detectorpars(\image_s)$} is already an ``exponentiated'' signal, we do a weighted normalization without an exponential,
\textit{i.e.} {\small ${\bh'} = \sum_s\tilde{\bh}_s(\tilde{\bh}_{s}/\sum_{s'} (\tilde{\bh}_{s'} + \epsilon))$}, where $\epsilon=10^{-6}$ is added to prevent division by zero.

Note that differently from LF-Net~\cite{Ono18}, we do not perform a softmax along the scale dimension.
The scale-wise softmax in LF-Net is problematic as the computation for a softmax function relies on the input to the softmax being \emph{unbounded}.
For example, in order for the softmax function to behave as a max function, due to exponentiation, it is necessary that one of the input value reaches infinity~(i.e. the value that will correspond to the max), or that all other values to reach negative infinity.
However, at the network stage where softmax is applied in~\cite{Ono18}, the score range from zero to one, effectively making the softmax behave similarly to averaging.
Our formulation does not suffer from this drawback.

\paragraph{Backbone-based heatmap network.}
For experiments on natural images, we restrict the detector to only localize the objects without estimating object scales 
to simplify the task.
Therefore, we only use a single-scale heatmap for this setting.
Also, because the number of images in our
dataset is limited, we leverage a pretrained ResNet34~\cite{He15} as the backbone feature extractor.
Specifically, we resize the input images to have a shorter edge of 224 pixels and use the output of fourth convolution block -- a tensor of $H/16 \times W/16 \times 256$ where $H$ and $W$ are the height and the width of the input image, respectively, and 256 is the number of channels.
We further append a $1\times1$ conv layers to reduce the heatmap to have 3 channels: response, $x$-offset, and $y$-offset.
We make use of offsets since
our heatmap size is only $1/16$ of the input image size, 
and
integer pixel coordinates on the heatmap cannot provide accurate 
localization
on the input image.
We also do not use a \textit{local spatial softmax} operator for this setting due to small heatmap size.
While not using spatial softmax makes our heatmaps similar to the ones in~\citeA{katharopoulos2019processing}, we note that we still rely on top-K, rather than the sampling-without-replacement approach of~\citeA{katharopoulos2019processing}, and is therefore easy to expand to various K.

\section{Generative model for the heatmap}
\label{sec:inversemap}

\vspace{-1em}
\paragraph{Standard architecture.}
To convert optimized locations into ideal heatmaps (see \Section{architecture}), 
as our standard architecture
we apply a simple model where the heatmap is zero everywhere else except on the corresponding keypoint locations (patch center).
However, as the optimized patch parameters 
can be floats corresponding to subpixel locations,
we need to quantize them with care
to turn them back into a heatmap
.
For the spatial locations we simply round to the nearest pixel, which at most creates a quantization error of half a pixel, which does not cause problems in practice.
For scale however, simple nearest-neighbor assignment causes too much quantization error as our scale-space is sparsely sampled.
We therefore assign values to the two nearest neighboring scales in a way that the center of mass would be the optimized scale value, making sure $\parameters = \extractor(\splatter(\parameters))$.

\paragraph{Backbone-based heatmap network.}
For the backbone-based network, as the feature map is very coarse, we found using a single pixel insufficient. Hence,
we use a Gaussian kernel to reconstruct the first channel (response channel) of the ideal heatmap from the optimized keypoints:
\begin{align}
    H_{\mathbf{p}}^{r} = \ \sum\limits_{i=1}^k &\exp(-\frac{1}{2 }(\lfloor\mathbf{x}_i\rfloor-\mathbf{p})^T\Sigma^{-1}(\lfloor\mathbf{x}_i\rfloor-\mathbf{p}))\nonumber
    \;,
\label{eq:gau_kernal}
\end{align}
where $\lfloor\mathbf{x}_i\rfloor$ are the rounded
optimized keypoints, $\mathbf{p}$ is a pixel location on heatmap.
We use one eighth of the patch size as the values of the diagonal covariance matrix $\Sigma$.

For the $x$ and $y$ offset channels we only supervise pixels which contain optimized keypoints:
\begin{align}
    H_{\mathbf{p}}^{xy} = \mathbf{x}_i - \lfloor\mathbf{x}_i\rfloor \quad \mbox{ if } \mathbf{p}=\lfloor\mathbf{x}_i\rfloor_i\in\mathbf{x}
    \;.
\end{align}
To train the detector, we use the $\ell_2$ loss for response channel as in \Eq{phase2}, and for the offset channels we use a Huber loss~\citeA{Huber64} as in~\citeA{Girshick15}:
\begin{align}
     \text{loss} =
     \begin{cases}
        0.5x^2 &\text{if $\lvert x \rvert<1$}\\
        \lvert x \rvert-0.5 & \text{otherwise}
     \end{cases}
     \;.
\label{eq:smooth_l1}
\end{align}

\section{Additional implementation details}

\subsection{Task-specific networks}
\label{sec:implementation}

\vspace{-1em}
\paragraph{MIST auto-encoder network.}
The input layer of the autoencoder is $32 \times 32 \times C$ where $C$ is the number of color channels.
We use five up/down-sampling levels.
Each level is made of three standard non-bottleneck ResNet v1 blocks \cite{He15}
and each ResNet block uses a number of channels that doubles after each downsampling step.
ResNet blocks use $3\times3$ convolutions of stride 1 with ReLU activation.
For downsampling we use 2D max pooling with $2 \times 2$ stride and kernel.
For upsampling we use 2D transposed convolutions with $2\times 2$ stride and kernel.
The output layer uses a sigmoid function, and we use layer normalization before each convolution layer.

\paragraph{MIST classification network for MNIST dataset.}
We use the same architecture as the encoder part of the auto-encoder and append a dense layer to it to map the latent space to the score vector of our 10 digit classes.

\paragraph{MIST classification network for PASCAL+COCO dataset.}
We crop patches at keypoint locations on \emph{the feature map} from fourth convolution block and feed the patches into a single ResNet block -- same as the one used for the auto-encoder network -- followed by a global average pooling layer, and a dense layer that converts the output into 21 logit values for classification.

\subsection{Implementations of compared methods}

\vspace{-1em}
\paragraph{Baseline unsupervised reconstruction methods.}
\label{sec:baseline_details}
To implement the {\it Esl16}~\cite{Eslami16} baseline, we use a publicly available third-party implementation\footnote{\url{https://github.com/aakhundov/tf-attend-infer-repeat}}.
We note that their method was originally applied to a dataset consisting of images of 0, 1, or 2 digits with equal probability.
We found that the model failed to converge unless it was trained with examples where various number of total digits exist, so for fair comparison, we populate the training set with images consisting of all numbers of digits between 0 and 9. 
For the {\it Zha18}~\cite{Zhang18b} baseline, we use the authors' implementation and hyperparameters. 

\paragraph{Differentiable top-K (\emph{Xie20})~\cite{Xie20}.}
We implemented their method as a PyTorch module according to the pseudo-code provided in \cite{Xie20}.
In addition to the provided pseudo-code, we add a small epsilon clipping for the division operations within the equations for stability.
The differentiable top-K operation in~\cite{Xie20}
outputs a top-k selection mask $\mathbf{m}\in (0,1)^N$, where $N$ is number of elements to select from
-- 
top-K elements have a mask score close to 1, and non selected elements have a mask score close to 0.
To apply this method to our task, one needs to then apply this mask to 
the classification results of patches at all heatmap location, as the operation is not indexing, but rather masking -- 
this is how differentiability is obtained in this method.
It is therefore necessary that all results that gets masks to stay in memory, and requires a smaller heatmap to be trained on reasonable system -- we use a 
\emph{GeForce RTX 2080 Ti} with 11GB memory.
In addition, we modify the
classification loss in Eq.\eq{mil_easy} to incorporate the selection mask:
\begin{equation}
    \loss_\mathrm{task} = \left\|
        \frac{1}{L}\sum_{l=1}^{L}\by_l - \frac{1}{K}\hat{{\bp}}\times \mathbf{m}
    \right\|_2^2
    \;,
\label{eq:task_diff_top_k}
\end{equation}
where $L$ is the number of instances in the image, $\hat{\bp}$ is the $C\times N$ classification score matrix, $\mathbf{m}$ is the mask vector of size $N$, with $C$ being the number of classes.
Note that the only difference here is that the top-K selection via indexing in the main paper has know been transformed into a mask-based selection.

\section{Non-uniform distributions}
\begin{figure}[ht]
\centering
\includegraphics[width=\linewidth]{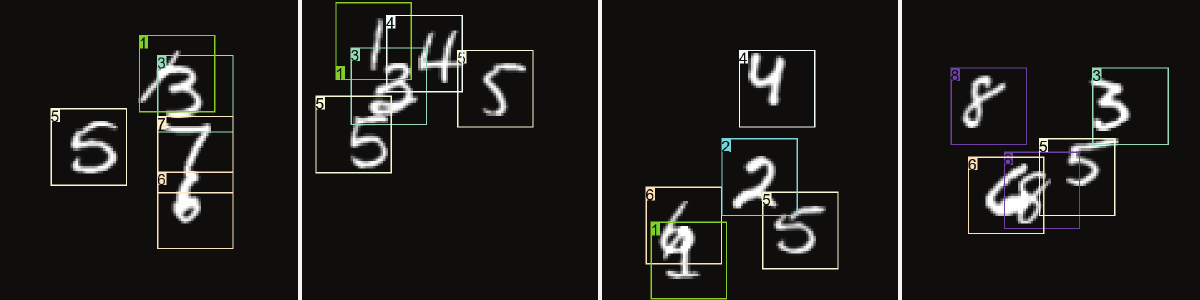}
\caption{Examples with uneven distributions of digits.}
\label{fig:nonuniform}
\vspace{-0.5em}
\end{figure}

Although the images we show in \Figure{mnist_gen} involve small displacements from a uniformly spaced grid, our method does not require the keypoints to be evenly spread. 
As shown in \Figure{nonuniform}, our method is able to successfully learn even when the digits are placed unevenly.
Note that, as our detector is fully convolutional and local, it does not learn the absolute location of keypoints.
In fact, we weakened the randomness of the locations for fairness against \cite{Zhang18b}, which is not designed to deal with severe displacements.

{\small
\bibliographystyleA{ieee_fullname}
\bibliographyA{string,vision,learning,add}
}

\end{document}